# Evolutionary computing-based image segmentation method to detect defects and features in Additive Friction Stir Deposition Process

AKSHANSH MISHRA[1,2], EYOB MESSELE SEFENE[3,4,] SHIVRAMAN THAPLIYAL[5,*]

1-School of Industrial and Information Engineering, Politecnico Di Milano, Milan, Italy
2-Computational Materials Research Group, AI Fab Lab, Uttar Pradesh, India
3-Department of Mechanical Engineering, National Taiwan University of Science and Technology, Taipei, Taiwan
4-Faculty of Mechanical and Industrial Engineering, Bahir Dar Institute of Technology, Bahir Dar University, P.O. Box 26, Bahir Dar, Ethiopia
5-Department of Mechanical Engineering, NIT Warangal, Telangana 506004, India

Corresponding author mail id: shivraman@nitw.ac.in

This work proposes an evolutionary computing-based image segmentation approach for analyzing soundness in Additive Friction Stir Deposition (AFSD) processes. Particle Swarm Optimization (PSO) was employed to determine optimal segmentation thresholds for detecting defects and features in multilayer AFSD builds. The methodology integrates gradient magnitude analysis with distance transforms to create novel attention-weighted visualizations that highlight critical interface regions. Five AFSD samples processed under different conditions were analyzed using multiple visualization techniques i.e. self-attention maps, and multi-channel visualization. These complementary approaches reveal subtle material transition zones and potential defect regions which were not readily observable through conventional imaging. The PSO algorithm automatically identified optimal threshold values (ranging from 156-173) for each sample, enabling precise segmentation of material interfaces. The multi-channel visualization technique effectively combines boundary information (red channel), spatial relationships (green channel), and material density data (blue channel) into cohesive representations that quantify interface quality. The results demonstrate that attention-based analysis successfully identifies regions of incomplete bonding and inhomogeneities in AFSD joints, providing quantitative metrics for process optimization and quality assessment of additively manufactured components.

## INTRODUCTION

In past few decades, the metal based additive manufacturing (MBAM) processes has shifted the paradigm in manufacturing by altering the design, production and distribution of the products. The design flexibility, ease of handling complexity and sustainability offered by MBAM processes resulted in adoption of this process in various engineering applications. However, the product development by AM processes encounters challenges i.e. technical, operational and economical and addressing these challenges is the focus of research for the scientific community. The technical challenges/defects i.e. porosity, delamination, balling, and surface roughness results in premature failure of AM components. Therefore, the detection and identification of these defect is important for process optimization and optimization. Image segmentation plays a crucial role in enhancing defect detection across various industrial and engineering applications. This advanced computer vision technique divides digital images into multiple segments or regions, each corresponding to a distinct object or part [1-4]. When applied to defect detection, image segmentation offers numerous advantages that significantly improve the accuracy and efficiency of quality control processes. The image segmentation enables able to pinpoint the region of interest with the image which offers a focused and accurate defect analysis. Additionally, the process of object or surface separation from backdrop and other extraneous aspects, the system can focus its processing capacity on the most important areas [5-8]. The detection system can identify differences in texture, color, shape, or other pertinent characteristics that might point to the existence of a problem thanks to this granular analysis. Even in intricate or visually busy settings, the system can more quickly detect anomalies or departures from typical patterns by looking at each of these split zones separately [9-13]. The image segmentation system can apply various detection algorithms or criteria to each section of an image depending on its unique properties by segmenting the image into separate parts. This flexibility allows image segmentation techniques application in various manufacturing process i.e. welding, machining and rolling [14-16]. Considering the ease and success of image segmentation methods in various manufacturing processes these methods have been adopted for additive manufacturing process also. Wong et al. [17] addressed the challenge of segmenting internal defects in additive manufacturing (AM) components using X-ray Computed Tomography (XCT) images, a critical task for ensuring part reliability. While XCT is widely employed for non-destructive evaluation, traditional manual thresholding methods for defect detection are labor-intensive, subjective, and struggle with variability in defect morphology. The authors proposed a U-Net-based convolutional neural network (CNN) framework to automate porosity and crack segmentation, achieving exceptional accuracy with a mean intersection over union (IOU) of 0.993. Saini et al. [18] addressed the critical challenge of in situ quality assessment in additive manufacturing (AM) by evaluating traditional image segmentation methods for real-time defect detection and dimensional analysis. Their work focused on overcoming the limitations of post-print inspection techniques like XCT, which are unsuitable for real-time monitoring during fabrication. The study systematically compared five segmentation methods i.e. simple thresholding, adaptive thresholding, Sobel edge detection, Canny edge detection, and watershed transform using a custom-built vision system integrated into an open-frame Fused Filament Fabrication (FFF) printer. The experimental setup, powered by a Raspberry Pi and Python-based software, analyzed segmentation performance under varying contrast conditions, with metrics including accuracy, precision, recall, and the Jaccard index. The past literature revealed that the image segmentation for defect detection in additive manufacturing process resulted in acceptable performance. However, the research focusing the implementation these techniques for defect detection in solid state additive manufacturing techniques i.e. Additive Friction Stir Deposition (AFSD) is limited. The Additive Friction Stir Deposition (AFSD) process has shown significant potential in producing high-quality multi-layer builds [19-23]. However, a persistent challenge in this method is ensuring the integrity of bonding between layers, particularly in cases where deposition is done at room temperature without the aid of cooling plates. AFSD exhibits issues such as incomplete bonding and unfilled regions in the deposited layers which can compromise the mechanical performance and durability of the components. Defect detection and process optimization using conventional techniques encounter challenges like inaccuracy, high processing time and dependency on the expertise. Therefore, to address these challenges, there is a need for a more robust and accurate defect detection method. Hence, this study aims to develop an evolutionary computing-based image segmentation approach using Particle Swarm Optimization (PSO) to identify and evaluate defects, including lack of bonding and unfilled regions, in AFSD multi-layer builds.

## METHODOLOGY

This work utilizes multilayer builds of ultra-high strength steel (Aermet) on steel substrate performed by Chou et al. [24]. The AFSD process involved depositing multiple layers of material at room temperature, with and without a cooling plate, which in some cases resulted in unfilled regions and a lack of bonding between the layers. The image processing was performed in Python. The segmentation optimization was performed using particle swarm optimization (PSO) which is a is a population-based optimization technique that simulates the social behavior of particles to find optimal solutions. In this context, PSO was used to determine the best threshold value for segmenting the grayscale images. A fitness function was defined to evaluate each threshold, which involved applying a binary threshold to the image and then measuring the effectiveness of segmentation based on edge detection. A penalty was introduced for cases where the segmentation resulted in predominantly black or white images, which would indicate poor segmentation, such as either over-segmentation or under-segmentation. The optimization process was run with the threshold bounds set between 50 and 200 to focus on reasonable segmentation values. The PSO algorithm iteratively refined the threshold values over multiple generations (with a swarm size of 30 and a maximum of 100 iterations), searching for the value that minimized the sum of edges. Once the optimal threshold was identified, it was applied to the original grayscale images to generate segmented images that clearly delineated the deposited regions from defects such as unfilled areas and unbonded regions. Self-attention map-based image segmentation employs transformer architectures to analyze relationships between all regions of an image, enabling precise feature localization by prioritizing global context. Unlike traditional convolutional methods restricted to local patterns, self-attention mechanisms compute interactions between pixels or patches, producing maps that emphasize semantic connections, such as object boundaries or texture coherence. These maps dynamically weight pixel relevance based on holistic image context, improving segmentation accuracy in complex scenarios like medical imaging or satellite analysis.

## RESULTS AND DISCUSSION

*PSO based segmentation and gradient magnitude visualization*

Figure 1 depict the canny edge segmented AFSD deposit image obtained after swarm-based optimization. The segmented images were compared with the optical micrographs of the additive friction stir deposited (AFSed) fabricated under different condition. The edge detection converts the micrographs into black and white image to capture any sharp change in the image feature. This technique aid to identify the defects, gaps and improper consolidation but fails to identify distinct features like material flow and mixing.

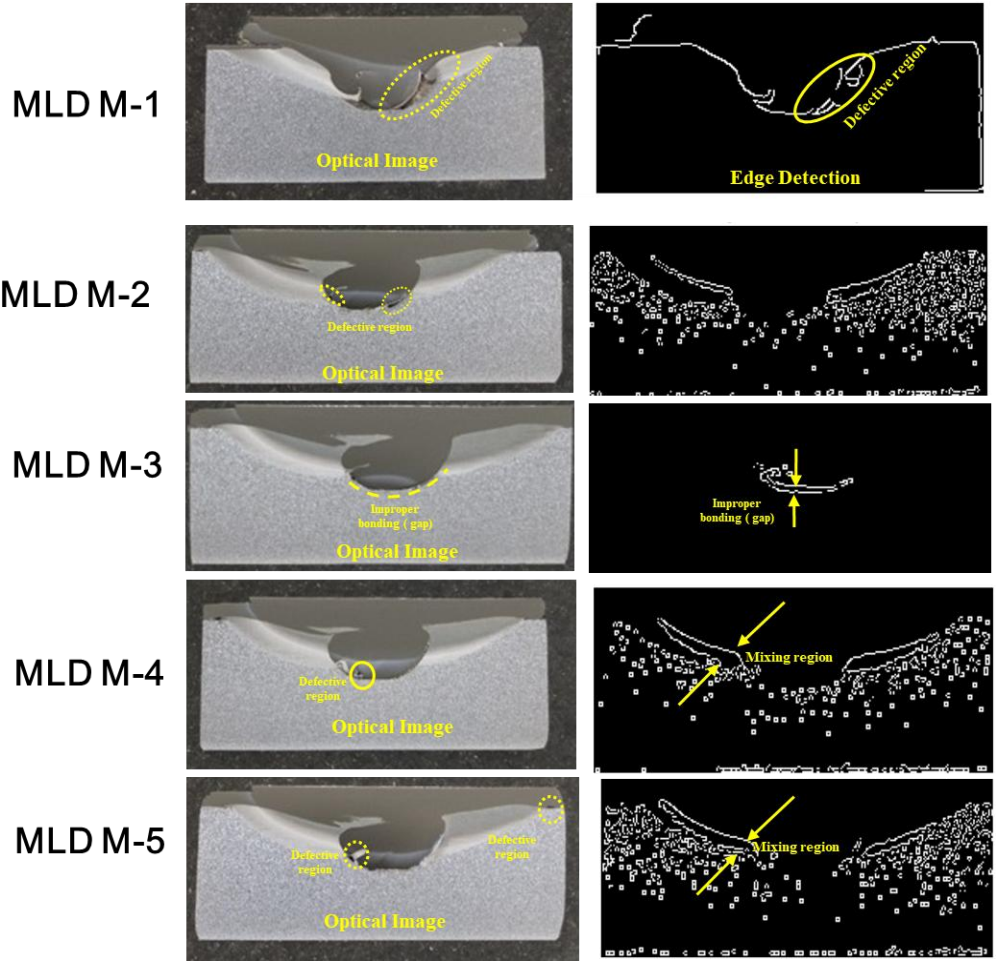

**Figure 1**. Image segmentation results for five samples. Each subfigure shows the original grayscale image (left) and the corresponding PSO-optimized segmented image (right), highlighting regions of lack of bonding and unfilled areas in the multi-layer builds deposited at room temperature

Figure 2 shows the multi-channel visualizations plots. The consistent colour mapping reveals important material characteristics and processing-related features. The cyan/light blue horizontal regions represent high attention areas where significant material property gradients exist, indicating the most critical interface zones where different materials or phases meet. These are surrounded by magenta/reddish boundaries that highlight the precise transition regions with high gradient values. The U-shaped depressions in each sample display subtle geometric variations, likely resulting from differing processing parameters or material compositions during the AFSD process. Sample a shows more pronounced edge features with brighter cyan regions along the upper interface, suggesting stronger material transitions or possibly greater mechanical mixing in this region. The lower portions of all samples display a consistent granular pattern in purple, representing the more homogeneous substrate material with relatively uniform properties. Samples exhibit progressive variations in interface geometry and boundary characteristics, particularly in the depth and profile of the U- shaped region. These visualizations effectively integrate boundary information (red channel), spatial relationships (green channel), and material density data (blue channel) into a cohesive representation that reveals subtle differences in bonding characteristics and material distribution that might not be apparent in conventional imaging techniques.

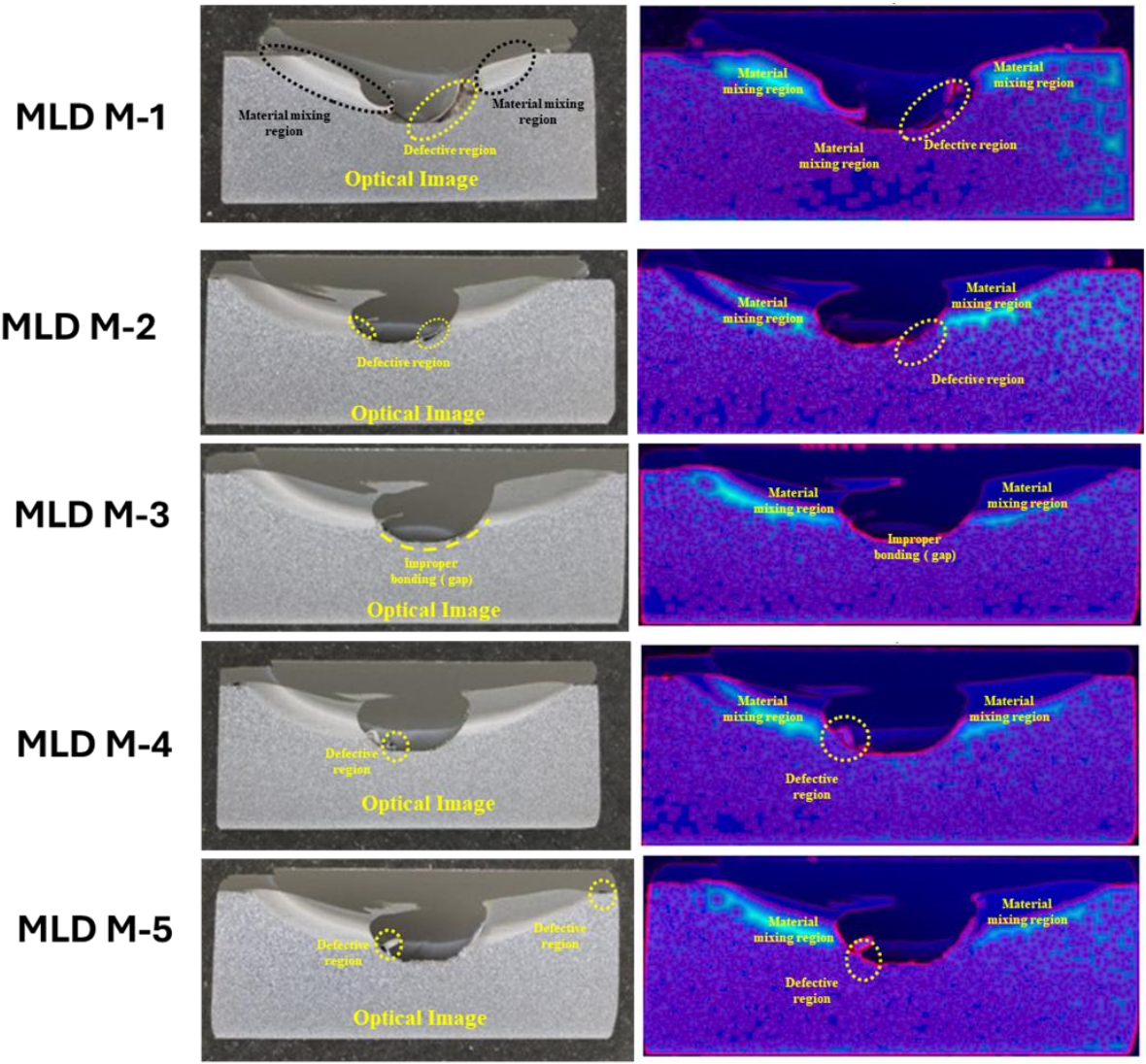

**Figure 2.** Multi-channel visualizations of AFSD interfaces across multiple samples (a-e), highlighting material transition zones. Color mapping represents a multi-channel composite where cyan/light blue regions indicate high-gradient areas associated with mechanical property

*Attention-Based Analysis of AFSD Material Interfaces*

The attention-based analysis of AFSD samples employs a multi-stage mathematical framework that highlights critical regions of interest while suppressing less informative areas (Appendix I). Figure 3 shows the attention overlay plots where each plot highlights regions of significant material transitions through color-coded attention values. MLD-M1 (a) shows a prominent U-shaped interface region with well-defined yellow boundaries indicating high gradient values. The upper rim exhibits particularly strong attention highlights (bright yellow green), suggesting pronounced material transitions. The cyan-colored lower regions indicate moderate attention areas with more uniform material distribution. Small dark blue patches in the substrate suggest localized variations in material homogeneity. MLD-M2 (b) displays a shallower U-shape with more gradual interface transitions. The yellow boundary is thinner but continuous, indicating consistent material interface without any defect (Figure 3b). Additionally, the cyan region is more extensive, suggesting a larger zone of moderate material gradients and the upper edge transitions appear slightly less intense than in MLD-M1. MLD-M3 features distinct dotted/segmented yellow highlights along the upper surface, indicating possible intermittent bonding or discrete interface features. The U-shaped depression shows sharp yellow boundaries at the bottom curve, suggesting high attention regions where material stress concentration likely occurs. MLD-M4 exhibits a more asymmetric interface with stronger yellow highlights on the right side of the U-shaped region. This suggests uneven material mixing or property distribution across the interface. The overall cyan region is like other samples, but shows more scattered dark blue patches, indicating greater heterogeneity in the substrate. MLD-M5 shows the most complex interface structure with multiple yellow highlight regions, particularly at the bottom of the U-shape where material interfaces appear most pronounced. The upper rim shows varying attention intensities, suggesting non-uniform material transition quality along the interface.

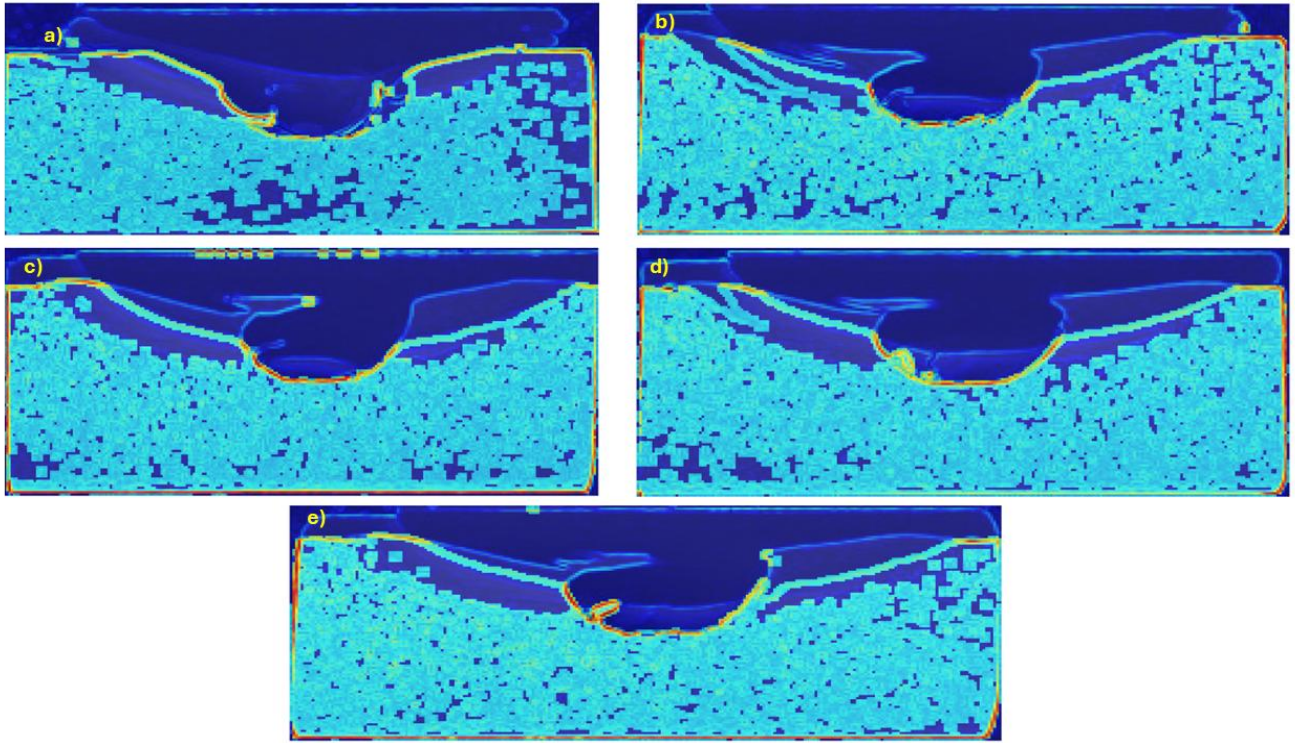

**Figure 3.** Attention overlay plots of five different Material Layer Deposition (MLD) samples (a-e) showing critical interface characteristics in Additive Friction Stir Deposition (AFSD) processing. Yellow/bright boundaries highlight regions of maximum gradient magnitude where significant material property transitions occur, cyan/light blue areas indicate moderate material property variations, and dark blue regions represent more homogeneous material zones. The visualization integrates edge-enhanced gradient information with distance transforms to reveal subtle processing-induced variations across samples MLD-M1 through MLD-M5, enabling quantitative comparison of interface quality and material distribution in AFSD joints.

Figure 4 shows the comparison between canny edge detection images and self-attention plots. The self-attention map shows pronounced red-orange boundaries along the upper interfaces with attention values of 0.6-0.8, indicating strong material transitions. The interface has a distinctive U-shape with particularly high attention at the corners where stress concentration is likely to occur. The bulk material below displays uniform light blue coloration (~0.2-0.3 attention value), suggesting consistent material properties. MLD M-2 (Threshold=166) exhibits a more pronounced white-to-red gradient along the upper interface with slightly higher segmentation threshold. The attention highlights are more focused at specific junction points rather than being continuously distributed, potentially indicating localized interface quality variations. The thin red-orange boundary line precisely tracks the material interface region where property gradients are highest. MLD M-3 (Threshold=167) continues this trend with similar threshold value but shows distinct differences in the attention distribution, particularly along the right side of the interface where attention values reach 0.6 (orange). This asymmetry in attention distribution suggests potential non-uniformity in the material bonding process across the sample width. MLD M-4 (Threshold=170) displays a more focused attention pattern with higher concentration at the bottom of the U-shaped interface where attention values peak in the red range (~0.6-0.7). The increased threshold results in cleaner segmentation with less noise, allowing the attention mechanism to more precisely identify critical regions. MLD M-5 (Threshold=173) shows the highest threshold value and consequently the most refined attention map. The interface boundary shows discrete high-attention spots rather than continuous lines, suggesting that at this threshold, the algorithm identifies specific points of interest where material property transitions are most significant.

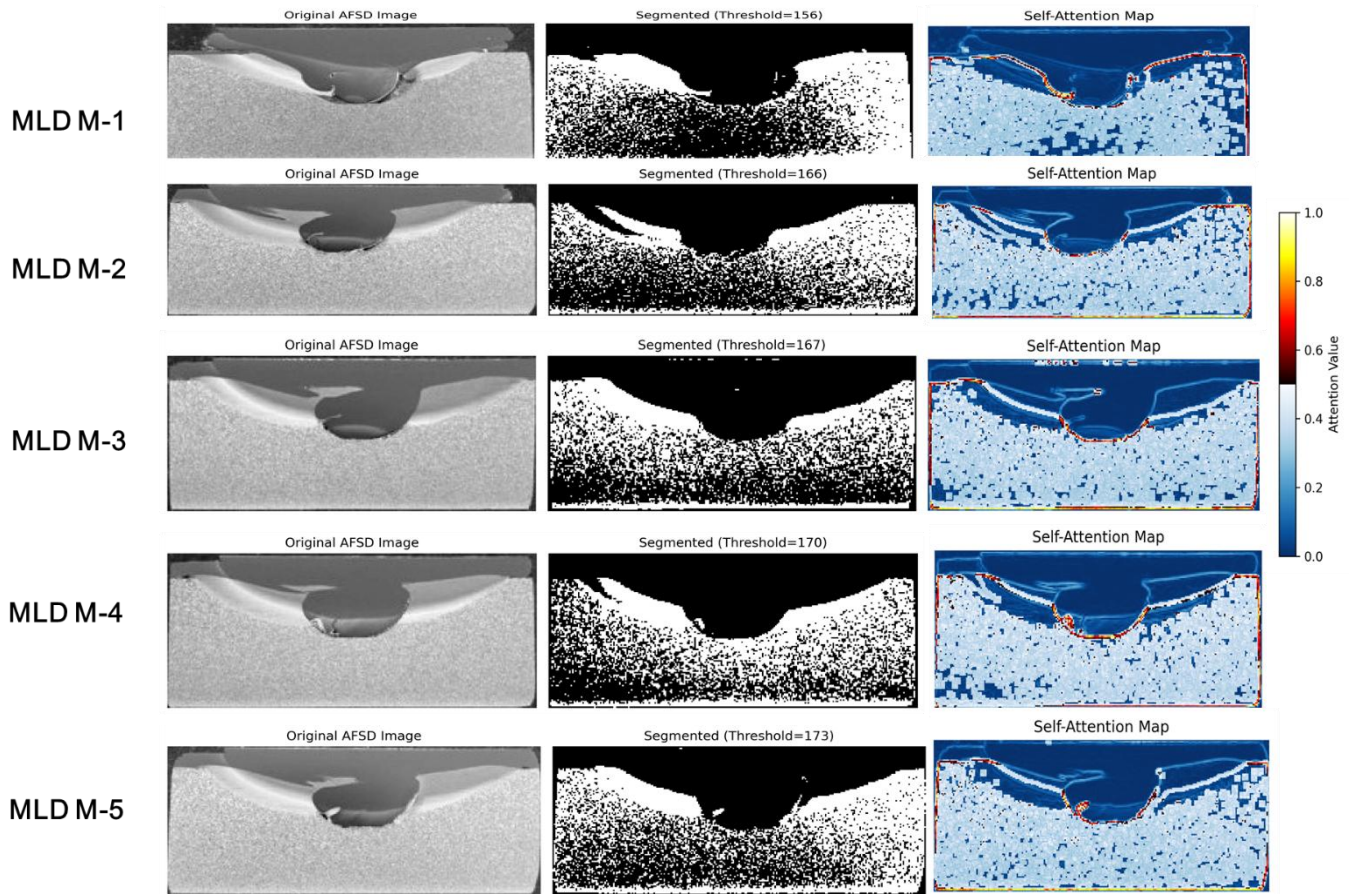

**Figure 4.** Visualization of material interface characteristics in AFSD samples through self-attention analysis. Each row presents the progression from original grayscale micrographs (left) through PSO-optimized binary segmentation (center) to self-attention maps (right). The automatically determined threshold values (156-173) adaptively segment each sample

## CONCLUSION

- A novel evolutionary computing framework integrating Particle Swarm Optimization (PSO) and attention-based visualization was developed to analyze material interfaces in Additive Friction Stir Deposition (AFSD), addressing adaptive segmentation and defect detection challenges in additive manufacturing.
- The PSO algorithm autonomously determined optimal segmentation thresholds, enabling precise interface identification despite variations in imaging conditions and sample heterogeneity.
- Advanced visualization techniques (gradient magnitude analysis, distance transforms, multi-channel attention mapping) revealed subtle defects (e.g., voids, incomplete diffusion zones) undetectable by conventional imaging methods.
- Quantitative correlations between processing parameters (temperature, deposition speed) and bond integrity were established, yielding predictive metrics for interface quality (transition sharpness, defect density).
- The framework enables non-destructive quality assessment in AFSD manufacturing, reducing reliance on destructive testing and providing actionable feedback for process optimization.
- Future work should validate derived metrics against mechanical performance data (tensile strength, fatigue resistance) and expand the methodology to multi-material AFSD systems.
- Integration of machine learning for real-time defect classification and parameter optimization could enhance industrial applicability, advancing intelligent process control in next-generation AFSD systems.